\documentclass[lettersize,journal,review]{IEEEtran}
\usepackage{amsmath,amsfonts}
\usepackage{algorithmic}
\usepackage{algorithm}
\usepackage{array}
\usepackage[caption=false,font=normalsize,labelfont=sf,textfont=sf]{subfig}
\usepackage{textcomp}
\usepackage{stfloats}
\usepackage{url}
\usepackage{verbatim}
\usepackage{graphicx}
\usepackage{cite}
\usepackage{multirow}
\usepackage{booktabs}
\usepackage{marvosym}
\usepackage{pifont}
\usepackage{amsmath}
\usepackage{xcolor}

\begin{document}

\title{Selective Volume Mixup for Video \\ Action Recognition}

\author{
Yi Tan, Zhaofan Qiu, Yanbin Hao,~\IEEEmembership{Member,~IEEE,} Ting Yao,~\IEEEmembership{Member,~IEEE,} and Tao Mei,~\IEEEmembership{Fellow,~IEEE}
\thanks{Y. Tan and Y. Hao are with the School of Information Science and Technology, University of Science and Technology of China, Anhui, 230009, China. E-mail: ty133@mail.ustc.edu.cn, haoyanbin@hotmail.com.}

\thanks{Z. Qiu, T. Yao and T. Mei are with HiDream.ai, Beijing, 100000, China. E-mail: zhaofanqiu@gmail.com ,tingyao.ustc@gmail.com, tmei@live.com.}

\thanks{Y. Hao is the corresponding author.}

}

\markboth{Journal of \LaTeX\ Class Files,~Vol.~14, No.~8, August~2021}%
{Shell \MakeLowercase{\textit{et al.}}: A Sample Article Using IEEEtran.cls for IEEE Journals}

\IEEEpubid{0000--0000/00\$00.00~\copyright~2021 IEEE}

\maketitle
\begin{abstract}
The recent advances in Convolutional Neural Networks (CNNs) and Vision Transformers have convincingly demonstrated high learning capability for video action recognition on large datasets. Nevertheless, deep models often suffer from the overfitting effect on small-scale datasets with a limited number of training videos. A common solution is to exploit the existing image augmentation strategies for each frame individually including Mixup, Cutmix, and RandAugment, which are not particularly optimized for video data. In this paper, we propose a novel video augmentation strategy named Selective Volume Mixup (SV-Mix) to improve the generalization ability of deep models with limited training videos. SV-Mix devises a learnable selective module to choose the most informative volumes from two videos and mixes the volumes up to achieve a new training video. Technically, we propose two new modules, i.e., a spatial selective module to select the local patches for each spatial position, and a temporal selective module to mix the entire frames for each timestamp and maintain the spatial pattern. At each time, we randomly choose one of the two modules to expand the diversity of training samples. The selective modules are jointly optimized with the video action recognition framework to find the optimal augmentation strategy. We empirically demonstrate the merits of the SV-Mix augmentation on a wide range of video action recognition benchmarks and consistently boot the performances of both CNN-based and transformer-based models. The code is available at \url{https://github.com/ty-97/Seletive-Volume-Mix}
\end{abstract}


\section{Introduction}
\IEEEPARstart{D}{eep} models, including CNN and transformer-based architectures, have successfully proven highly effective for understanding multimedia content on large-scale datasets. To date in the literature, there are various large models that push the limits of multimedia analysis systems, e.g., Vision Transformer \cite{dosovitskiyimage}, Swin Transformer \cite{liu2021swin}, ConvNeXt \cite{liu2022convnet} for image classification, MViT \cite{li2022mvitv2}, Video Swin \cite{liu2021video}, Uniformer \cite{li2021uniformer} for video analysis, CLIP \cite{radford2021learning}, GLIP \cite{li2022grounded} for cross-modality understanding. Nevertheless, the impressive performances of these models highly rely on large-scale datasets and are easily affected by the overfitting effect on the tasks with insufficient training data. Such an issue becomes even worse particularly for video action recognition due to the difficulty of achieving large amounts of video data and expensive efforts for labeling.

\begin{figure}[t]
  \centering
  \includegraphics[width=0.9\linewidth]{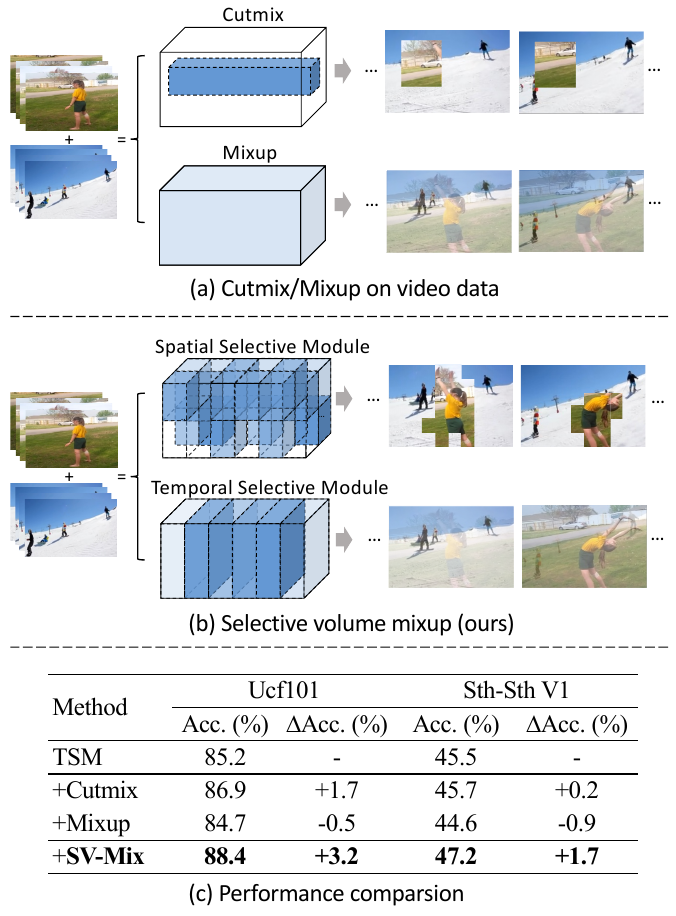}
    \vspace{-0.3cm}
  \caption{The intuition of (a) the typical Cutmix \cite{yun2019cutmix} and Mixup \cite{zhangmixup} augmentations on video data, and (b) our Selective Volume Mixup (SV-Mix). The typical methods randomly combine regions or entire frames from two videos and may lose crucial information. In contrast, our SV-Mix contains learnable selective modules to adaptively select valuable volumes. A tapas performance comparison between Cutmix/Mixup and our SV-Mix is also shown in (c).}
  \label{fig:instance}
  \vspace{-0.2cm}
\end{figure}

To alleviate this issue, a general practice is to exploit network regularization and data augmentation to preserve the effectiveness of large models with limited training data. Among these strategies, the network regularization methods, including dropout \cite{srivastava2014dropout}, drop path \cite{huang2016deep}, and weight decay \cite{loshchilov2017decoupled}, are general network training schemes across different tasks and can be directly utilized for video models. However, the policy of data augmentation should be specially designed for different data formats since it is highly related to the intrinsic properties of the input modality. The current standard data augmentation strategy for \IEEEpubidadjcol video data (e.g., in \cite{feichtenhofer2019slowfast,liu2021video,li2022mvitv2}) is to simply perform the existing image augmentation to each frame individually, as illustrated in Figure \ref{fig:instance}(a). This solution is straightforward but ignores the properties of video data, e.g., the temporal correlation across frames, and thus weakens the effectiveness of data augmentation. Moreover, these strategies are all manually devised and not learnable for different architectures/datasets, which requires more significant engineering effort of human experts to tune the hyper-parameters given a new architecture/dataset.

In this work, we aim at investigating a learnable data augmentation mechanism to facilitate the data efficiency for video action recognition. We start from the basic idea of the popular image augmentations, i.e., Mixup \cite{zhangmixup} and Cutmix \cite{yun2019cutmix}, that combines the content of two videos to obtain a new training video. Mixup blends the two training images by weighted summation, and Cutmix randomly exchanges a local region of two samples. These two manually designed strategies consistently show good performances in image models but are not acclimatized to the video domain. We speculate that the difficulty of video Mixup/Cutmix mainly originates from two aspects: 1) Video is an information-intensive media, and the labeled actions are related to objects, interaction, scene, etc. Randomly exchanging a local region (like in Cutmix) may lose some crucial information. 2) Video usually contains several background frames that do not contain the labeled action. Blending the background frame to the other video (like in Mixup) may dilute the useful cues of the original video. That motivates us to abandon the random blending approaches and devise a selective module to preserve the informative volumes in the mixing process.

To this end, we present a new Selective Volume Mixup (SV-Mix), as shown in Figure \ref{fig:instance}(b). Specifically, SV-Mix contains two modules, i.e., a spatial selective module and a temporal selective module. The former builds cross attention between the local patches with the same timestamp from two videos, respectively, to determine the preserved patches in each spatial position. The latter utilizes a similar attention mechanism but on the frame level to blend the most informative frames while keeping the spatial structure. The two selective modules are complementary to each other that select the mixed volumes on the patch level and frame level, respectively. Hence, we stochastically choose one of the two modules at each time to expand the divergence of the augmented samples. SV-Mix is jointly optimized with the action recognition framework in an end-to-end manner. Moreover, to avoid the bad influence between the augmentation network and the action recognition network, we devise a disentangled training pipeline, which exploits a slow-moving average of action recognition parameters instead of the training one to guide the optimization of SV-Mix. As a result, the gradients of both components are disentangled and lead to the convergence of both optimal data augmentation and optimal action recognition framework.

To the best of our knowledge, our work is the first to devise an end-to-end learnable augmentation strategy for videos. The design also leads to the elegant view of how to adaptively mix two videos while maintaining valuable information. We uniquely formulate the problem as cross attention between the volumes of two videos and devise two selection modules by mixing along the spatial dimension and temporal dimension, respectively. Extensive experiments on five datasets demonstrate the effectiveness of our proposal, and with different action recognition frameworks including both CNN-based and transformer-based methods, our SV-Mix consistently improves the performances over other augmentation strategies.

\section{Related work}
We briefly group the related works into two categories: video action recognition and data augmentation strategies.

\textbf{Video action recognition.} With the prevalence of deep learning in multimedia analysis, the dominant paradigm in modern video action recognition is deep neural networks. The research of deep models for video action recognition has proceeded along three dimensions: 2D CNNs, 3D CNNs and video transformers. 2D CNNs \cite{simonyan2014two,feichtenhofer2016convolutional,wang2020makes,diba2017deep,qiu2017deep} often treat a video as a sequence of frames or optical flow images, and directly extend the 2D image CNNs  for frame-level recognition. For instance, the famous two-stream networks proposed in \cite{simonyan2014two} apply two 2D CNNs separately on visual frames and stacked optical flow images. Later, Wang \emph{et al.} \cite{wang2016temporal} propose Temporal Segment Networks, which divide input video into several segments and sample one frame/optical flow image from each segment as the input of two-stream networks. Two-stream architecture is further extended by advanced fusion strategies \cite{feichtenhofer2016convolutional,wang2020makes}, feature encoding mechanism \cite{diba2017deep,qiu2017deep} and training process optimization \cite{huang2018toward}. Besides, spatio-temporal attention \cite{yang2020sta,hao2022spatio,tan2021selective} also demonstrates effectiveness to equip 2D CNNs with video dynamics modeling capability.

The above 2D CNNs proceed each frame individually at early layers, and the pixel-level temporal evolution across consecutive frames is seldom explored. To alleviate this issue, 3D CNNs \cite{ji20123d,tran2015learning,qiu2017learning,xie2018rethinking,tran2018closer,tran2019video,feichtenhofer2020x3d,varol2018long,diba2018temporal,hussein2019timeception,feichtenhofer2019slowfast,wang2018non,qiu2019learning,qiu2021optimization} are devised to directly learn spatio-temporal correlation from video clips via 3D convolution. A prototype of 3D CNNs is introduced in \cite{ji20123d} by replacing 2D convolution in 2D CNNs with 3D convolution. A widely adopted 3D CNN, called C3D \cite{tran2015learning}, is devised by expanding VGG-style 2D CNN to 3D manner with both 3D convolutions and 3D poolings. To reduce the expensive computations and the model size of 3D CNNs, the fully 3D convolution is decomposed into a spatial convolution plus a temporal convolution \cite{qiu2017learning,xie2018rethinking,tran2018closer,tan2022hierarchical} or a depth-wise convolution plus a point-wise convolution \cite{tran2019video,feichtenhofer2020x3d}. Another scheme to improve 3D CNNs is to expand the temporal receptive field. Varol \emph{et al.} present LTC architecture \cite{varol2018long} that increases the length of input clips while reducing the resolution of the input frame.
Furthermore, 3D convolution on different time scales \cite{diba2018temporal,hussein2019timeception,feichtenhofer2019slowfast} and holistic view of video \cite{wang2018non,qiu2019learning} are also proven to be effective on long-term modeling.

More recently, video transformers \cite{bertasius2021space,arnab2021vivit,fan2021multiscale,liu2021video,yan2022multiview,long2022stand,long2022dynamic,yang2022recurring,patrick2021keeping,li2021uniformer} become formidable competitors to 3D CNNs. The early works for video transformers, i.e., TimesSformer \cite{bertasius2021space} and ViViT \cite{arnab2021vivit} study the basic designs of video transformer including tubelet embedding and attention decomposition. MViT \cite{fan2021multiscale} and Video Swin \cite{liu2021video} follow the philosophy of CNNs, where the channel dimension increases while the spatial resolution shrinks with the layer going deeper, to reduce the computational cost. More fine-grained designs are proposed recently to improve video transformers, including Multiview Transformer \cite{yan2022multiview}, cross-frame attention \cite{long2022stand}, recurrent attention \cite{yang2022recurring}, trajectory attention \cite{patrick2021keeping}, and combining 3D convolutions \cite{li2021uniformer}.

\begin{figure*}[t]
  \centering
  \includegraphics[width=0.87\linewidth]{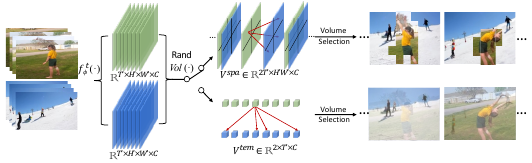}
    \vspace{-0.2cm}
  \caption{The overview of our Selective Volume Mixup (SV-Mix) data augmentation. Given two training videos, we first extract the volume-level feature map 
  with shape $\mathbb{R}^{T'\times H'\times W'\times C}$ for each video by an encoder $f_\phi^t(\cdot)$. 
  Next, we randomly instantiate a specific volume partition function $Vol(\cdot)$ to partition the feature map as $V^{spa}$ or $V^{tem}$ which respectively corresponding to patch level selection (spatial selective module) and frame level selection (temporal selective module).
  The selected volumes are then combined together to achieve an augmented video as the input of the subsequent action recognition framework.}
  \label{fig:framework}
  \vspace{-0.4cm}
\end{figure*}

\textbf{Data augmentation strategies.} The data augmentation strategy, as an important facility to alleviate the overfitting effect, has attracted intensive research interests in recent years. Take the augmentation strategies in the image domain as an example, Autoaugment \cite{cubuk2018autoaugment} formulates the augmentation process as the combination of sequential augmentation operations and proposes a search-based algorithm to tune the strength of each operation. Similarly, RandAugment \cite{cubuk2020randaugment} is also based on a group of augmentation operations but replaces the search process in auto augmentation by randomly choosing operations. In addition to the mentioned methods that apply global transformations to images, augmentations based on local deletion and sample mixing have also gained attention \cite{naveed2024survey}. Random erasing \cite{zhong2020random} randomly choose a local region of an image and erase the content inside the bounding box. Mixup \cite{zhangmixup} proposes to blend two samples by random weights and perform multi-label classification on the mixed image. Similarly, Cutmix \cite{yun2019cutmix} devises a strategy to exchange the pixels in a local region of two images to be mixed. Moreover, the such idea of mixing two images is improved by the advanced techniques including Saliency Grafting \cite{park2022saliency}, TransMix \cite{chen2022transmix} and Automix \cite{liu2022automix}. The characteristics of data augmentations (e.g. similarity and diversity) are also explored by Yang et al. \cite{yang2024investigating}.
Besides pure vision application, Mixgen \cite{hao2023mixgen} expands Mixup for multi-modal data augmentation by concatenate the text attributes.


More closely related to our work, VideoMix \cite{yun2020videomix} and DynaAugment \cite{kim2022exploring} remould the existing Cutmix and RandAugment to the video domain, respectively. The distinction between our work and these methods lies in our work aims to devise a learnable data augmentation strategy for action recognition. Learn2augment \cite{gowda2022learn2augment} proposes a learning-based approach to fuse foreground and background from different videos to achieve video augmentation. However, it requires additional policy optimization and pretrained segmentation models, whereas our approach offers a convenient end-to-end solution without the need for extra models or policy optimization.


\section{METHODOLOGY}
In this section, we deliberate our proposed Selective Volume Mixup (SV-Mix) for video action recognition. First, we briefly summarize the preliminaries of video model training using mixed video samples. Then, we detail the architecture of volume selection, and show how to utilize this architecture to construct spatial selective module and temporal selective module. Finally, a novel disentangled training pipeline is proposed to jointly optimize the volume selection modules and the action recognition framework. Figure \ref{fig:framework} illustrates the overview of our proposed SV-Mix.

\subsection{Preliminaries}
Given a video sample $\textit{x}\in \mathbb{R}^{T\times H\times W\times 3}$ which contains $T$ frames with size $H\times W$ and $3$ channels, the goal of the video action recognition model is to inference its one-hot class label $y \in \textbf{Y} = \{0,1\}^K$, where $K$ denotes the number of categories. In the action recognition pipeline with mixed video samples, the input video data and the corresponding label are rearranged as the linear interpolation of two or more videos, and in following statement, we focus on video mixing using two samples for conciseness. Particularly, given two video-label pairs $(\textit{x}_i, \textit{y}_i)$ and $(\textit{x}_j, \textit{y}_j)$, the mixed process of samples and labels can be represented as: 
\begin{align}
\widetilde{x} &= \textbf{M} \odot \textit{x}_i + (1-\textbf{M})\odot \textit{x}_j,
\\
\widetilde{y} &= \lambda \textit{y}_i + (1-\lambda) \textit{y}_j,
\end{align}
where $\odot$ denote hadamard product and $\textbf{M}\in [0,1]^{T\times H \times W}$ (omit channel for conciseness) is the mixing weights. Each element $\textbf{M}_{t,w,h}$ represents the mix proportion of a specific pixel. $\lambda\sim Beta(\alpha,\alpha)$ is the random label proportion. In addition, given $\lambda$, \textbf{M} satisfies $\frac{1}{THW}\sum_t^T\sum_h^H\sum_w^W = \lambda$. The goal of video model training is to find a group of parameters $\phi$ of deep neural network $f_\phi$ which minimize the following loss function in mini-batch:

\begin{equation}
\begin{split}
    \hat{\phi}=\arg\min_\phi \mathcal{L}(\phi),\quad\mathcal{L}(\phi) = \mathbb{E}_{\mathcal{B}}[\mathcal{L}_{sce}(f_\phi(\widetilde{x}),\widetilde{y})],
\end{split}
\label{opt}
\end{equation}
where $\mathcal{B}$ is the mini-batch during training iteration, $\mathcal{L}_{sce}$ denotes the soft-targat cross entropy loss. We omit the expectation of the loss function in the following for concise.

\subsection{Selective Volume Mixup}
In the traditional mixing process, the mixing weights $\textbf{M}$ are usually randomly sampled from a manually designed principle, e.g., the frame-level random weights in Mixup, and the random rectangle with value one in Cutmix. These strategies are not learnable and ignore the content of input videos. In contrast, our goal is to parameterize the generation process of mixing weights $\textbf{M}$. In the other words, we try to devise another neural network to predict the probability of each volume in the mixed video coming from $\textit{x}_i$.

We achieve this goal by answering two core design questions: 1) How to model the relationship between volumes from two input videos to obtain a most informative mixed video; 
2) With the additional temporal dimension, how to avoid the intensive full spatio-temporal dependency modeling and further leverage this property of video data for a modality-specific sample mixing; 
For the first question, we propose an attention-based block, which treats each volume in $x_i$ as a query and the volumes in $x_j$ as keys and values to evaluate if this volume should be maintained in the mixed video. For the second one, we propose to decompose the full spatio-temporal relation, and at each time, only calculate the attention along the spatial dimension or temporal dimension probabilistically.
Toward these, we first reformulate the generation of the mixed sample as:


\begin{equation}
\begin{split}
    \mathcal{G}_{\theta} ({x}_i,{x}_j,\lambda) = \mathcal{M}_{\theta} (Vol(x_i),Vol(x_j),\lambda) \odot x_i + \\
    (1-\mathcal{M}_{\theta} (Vol(x_i),Vol(x_j),\lambda)) \odot x_j,
\end{split}
\end{equation}
in which ${\theta}$ parameterised volume selection function $\mathcal{M}_{\theta} (\cdot)$ and the volume partitioning function $Vol(\cdot)$ implements the attention decomposition. We then detail these two components one by one.

\begin{figure}[t]
  \centering
  \includegraphics[width=0.8\linewidth]{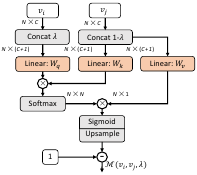}
  \caption{A diagram of the volume selection module in our SV-Mix. Given the volumes from two videos $v_i,v_j\in \mathbb{R}^{N\times C}$ where $N$ and $C$ are the number of volumes and channels, respectively, the attention weights across two videos are calculated through three trainable linear mappings $W_q,W_k,W_v$. Here $\lambda$ denotes the label proportion of the first video.}
  \label{fig:model}
  \vspace{-0.2cm}
\end{figure}

\textbf{Volume selection}.
Given video volume $v_i=Vol(x_i)$ and $v_j=Vol(x_j) \in \mathbb{R}^{N\times C}$, where $N$ and $C$ represent the number of volumes and channels under a specific $Vol(\cdot)$ function, we devise a cross-attention mechanism to model the relations between $v_i$ and $v_j$, as shown in Figure \ref{fig:model}. The attention response naturally serves as an importance measurement of each element, and thus we can easily transfer it to volume selection. Technically, the similarities between volumes is formulated as:
\begin{equation}
\begin{split}
    S(v_i,v_j,\lambda) = Softmax\left(\frac{(v_{i,\lambda}W_q) \otimes (v_{j,1-\lambda}W_k)^T}{\sqrt{d_k}}\right),
\end{split}
\end{equation}
where $W_q$ and $W_k \in \mathbb{R}^{(C+1)\times C}$ are learnable matrices for projecting volumes into queries and keys respectively. $d_k$ is the dimension of queries and keys. $v_{i,\lambda}$ and $v_{j,1-\lambda}$ denote $\lambda$ embedded volume representations achieved by concatenating $\lambda$ or $1-\lambda$ on the channel dimension. 
We then generate our selection weights $\mathcal{M}$ as the inverse of the attention response which is calculated as the summation of values using the similarity $S$ as weights:
\begin{equation}
\begin{split}
    \mathcal{M}(v_i,v_j,\lambda) = 1-Upsample\left( \delta \left( S(v_i,v_j,\lambda) \otimes  v_{j,1-\lambda}W_v\right) \right) ,
\end{split}
\end{equation}
where $Upsample(\cdot)$ infers the full spatio-temporal mixing weights from the $N$ sampled volumes, $\delta(\cdot)$ is the sigmoid operation to normalize the responds into $[0,1]$ as a proportion value and $W_v \in \mathbb{R}^{(C+1)\times 1}$ is the projection matrix for values. Please note that, here we utilize the inverse of attention responses as the mixing weights, since the higher attention responses usually indicate higher similarity with the other video but we prefer to preserve the most distinctive volumes.

\textbf{Volume partition}. Taking inspiration from the concept of spatio-temporal decomposition \cite{qiu2017learning, tran2018closer}, we employ a divide-and-conquer approach to partition the video into volumes from spatial or temporal perspective. 
This allows us to perform volume selection along a single dimension, thereby improving modeling efficiency and more importantly, enhancing the diversity of mixed video samples.
Notably, since the attention computation is shape agnostic for input data, individual spatial selection and temporal selection share the same parameters but partitioning the volumes along different dimensions.

Given the video feature $Z \in \mathbb{R}^{B \times T' \times H'  \times W' \times C}$ encoded by the backbone network $f_\phi^t(\cdot)$, we reshape $Z$ and achieve volumes as $V^{spa} \in \mathbb{R}^{BT' \times H'W' \times C}$ for spatial selection. Through moving the temporal dimension into the batch dimension, the attention is calculated along the spatial dimension for each timestamp individually.

Similarly, for the temporal selective module, we shrink the spatial dimension to gather volumes $V^{tem} \in \mathbb{R}^{B \times T' \times C}$
\begin{equation}
    V_{tem} = \frac{1}{H' \times W'} \sum_i^{H'} \sum_j^{W'} Z_{i,j}.
\end{equation}
With $V^{tem}$ as the input, the temporal selective module captures the relationship between frames instead of patches, and then assigns the same weight for all patches in the identical frame. Through temporal selection, we assign relative importance for each frame in the mixed video while maintaining the spatial pattern.

A common strategy to ensemble the spatial selective module and temporal selective module is to simply average the mixing weight from the two modules as
\begin{equation}
\label{eq:ae}
    \mathcal{M}_{en} = \frac{1}{2} \left(\mathcal{M}_{\theta}(v^{tem}_{i},v^{tem}_{j},\lambda)+\mathcal{M}_{\theta}(v^{spa}_{i},v^{spa}_{j},\lambda)\right).
\end{equation}
However, this strategy produces mixed training samples with only a single style, which limits the diversity \cite{gontijotradeoffs} of data augmentation in model training. Under this consideration, we propose to probabilistically ensemble spatial selective module and temporal selective module by randomly choosing one of the two modules at each time:
\begin{equation}
\label{eq:pe}
\mathcal{M}_{en} = 
\begin{cases}
\mathcal{M}_{\theta}(v^{tem}_{i},v^{tem}_{j},\lambda), & \mu \leq P \\ \mathcal{M}_{\theta}(v^{spa}_{i},v^{spa}_{j},\lambda), & \mu>P
\end{cases}
\end{equation}
where $\mu \sim U(0,1)$ is sampled from a uniform sampling, and $P$ is the switch probability between two modules. Here, we simply set $P=0.5$ to demonstrate the effectiveness of prebabilistically ensemble.

\textbf{Disentangled training pipeline}. By parameterising the sample mixing process, the forward propagation of the action recognition model changes from $f_\phi(x)$ to $f_\phi(\mathcal{G}_{\theta}(x_{i},x_{j},\lambda))$. An intuitive way to jointly optimize the selective modules $\theta$ and action recognition framework $\phi$ is to directly conduct end-to-end learning (we refer it to entangled training in the following parts) under the supervision of mixed label $\widetilde y  = \lambda y_i + (1-\lambda y_j)$. However, a gradient entanglement occurs in the optimization process of volume selective module $\theta$:



\begin{equation}
\frac{\partial \mathcal{L}_{sce}}{\partial \theta} = \textcolor{red}{\frac{\partial \mathcal{L}_{sce}(f_\phi (\mathcal{G}_\theta(x_{i},x_{j},\lambda)),\widetilde{y})}{\partial \mathcal{G}_\theta(x_{i},x_{j},\lambda)}} \cdot \textcolor{blue}{\frac{\partial \mathcal{G}_\theta(x_{i},x_{j},\lambda)}{\partial \theta}},
\end{equation}
which consists of two multiplicative components: one being the gradient of the loss function with respect to mixed sample $\mathcal{G}_\theta(x_{i},x_{j},\lambda)$ (marked by \textcolor{red}{red}), and the other being the gradient of $\mathcal{G}_\theta(x_{i},x_{j},\lambda)$ with respect to its parameters $\theta$ (marked by \textcolor{blue}{blue}). 
The first part of gradient heavily depends on the recognition network $f_\phi$. However, before convergence, the parameters $\phi$ of recognition network changes rapidly after each back-propagation step, leading to an unstable updating direction for the SV-Mix parameters $\theta$.
As a consequence, SV-Mix can only produce sub-optimal generated samples, further limiting the performance of the recognition model (shown in Figure \ref{fig:disen} and Table \ref{tab:en_vs_disen}).

\begin{figure}[t]
  \centering
  \includegraphics[width=\linewidth]{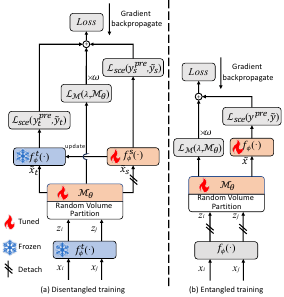}
  \caption{The proposed disentangled training pipeline for jointly optimizing SV-Mix and the action recognition networks $f_\phi^s$. In this pipeline, the gradient of SV-Mix is provided by a momentum-updated version of action recognition network $f_\phi^t$. Therefore, the gradients of SV-Mix and $f_\phi^s$ are disentangled, which stabilizes the training process.}
  \label{fig:pipline}
  \vspace{-0.5cm}
\end{figure}

To address the issue, we drew inspiration from BYOL \cite{grill2020bootstrap} whose training process is split into two branches: an online branch and a target branch, where the target branch is a moving average of the online branch, providing a stable supervision signal for the online one. Following this design, we decoupled the training of SV-Mix into two branches. In one branch, the recognition network is kept frozen and is set as the moving average of the unfrozen branch. In this setting, parameter $\phi$ in $\frac{\partial\mathcal{L}_{sce}\left(f_\phi\left(\mathcal{G}_\theta\left(x_i,x_j,\lambda\right)\right),\widetilde{y}\right)}{\partial\mathcal{G}_\theta\left(x_i,x_j,\lambda\right)}$ can be considered as a group of constant, ultimately leading to a stable $\frac{\partial \mathcal{L}_{sce}}{\partial \theta}$ for optimizing SV-Mix. Thus, the optimization process in one iteration is as demonstrated in Figure \ref{fig:pipline} (a): 1) frozen $f_{\phi}^t$ encodes two video samples $x_i$ and $x_j$ into semantic space $z_i$ and $z_j$; 2) randomly choose a volume partition strategy, e.g. $V^{spa}$ or $V^{tem}$, and generate two mixed video sample $\widetilde{x}_s$ and $\widetilde{x}_t$ for $f_{\phi}^s$ and $f_{\phi}^t$, respectively, to predict action categories; 3) update $\phi_s$ and $\theta$; 4) update $\phi_t$ via exponential moving average (EMA): $\phi_t \leftarrow m\phi_t+(1-m)\phi_s$, where $m \in [0,1)$ is the momentum coefficient.

In addition, we introduce an extra tributary loss for leading the mixing weight $\mathcal{M}_\theta$ to match the pre-sampled $\lambda$:

\begin{equation}
    \mathcal{L}_\mathcal{M}(\lambda, \mathcal{M}_\theta)=\| \lambda - \frac{1}{T\times H\times W}\sum_{i,j,k}\mathcal{M}_{\theta}^{i,j,k} \|
\end{equation}
We then scale $\mathcal{L}_\mathcal{M}$ by a coefficient $\omega$ and add it to the joint loss as:
\begin{equation}
    \mathcal{L}=\mathcal{L}_{sce}(y_t^{pre},\widetilde{y}_t) + \mathcal{L}_{sce}(y_s^{pre},\widetilde{y}_s)+\omega \mathcal{L}_\mathcal{M}(\lambda, \mathcal{M}_\theta).
\end{equation}




\section{Experiments}

In this section, we empirically evaluate the performance of the SV-Mix on various video recognition datasets, video recognition models and different settings to answer the following research questions:


\begin{itemize}[]
\item \textbf{RQ1}: How does SV-Mix perform on different video datasets when adopted on various recognition models?

\item \textbf{RQ2}: How does SV-Mix compare with other data augmentation methods?
\item \textbf{RQ3}: How do different settings affect SV-Mix?
\end{itemize}

\subsection{Datasets}
\textbf{Something-Something.} Something-Something dataset consists of 174 fine-grained action categories and 110k(V1) and 220k(V2) videos that depict humans performing everyday actions with common objects. Recognizing actions in the Something-Something dataset heavily relies on identifying key regions and frames. Classical random linear interpolation methods, such as Mixup \cite{zhangmixup} or Cutmix \cite{yun2019cutmix}, can weaken or delete these key patterns. 

\textbf{Mini-Kinetics.} Kinetics-400 \cite{carreira2017quo} is a widely used action recognition benchmark. It contains 240k training samples and 20k validation samples in 400 human action classes. Kinetics dataset mainly focus on static spatial appearance pattern. We create a mini version of Kinetics-400 dataset following \cite{chen2021deep} which accounts for half of the full Kinetics400 through randomly selecting half of the categories of Kinetics-400.

\textbf{Diving48}. Diving48 \cite{li2018resound} is a fine-grained video dataset comprising approximately 18,000 trimmed video clips of 48 unambiguous dive sequences in competitive diving. Due to the diverse sub-poses distributed throughout the timeline in the dive sequences, it is crucial to capture the local sub-poses over time. 

\textbf{EGTEA Gaze+}. EGTEA Gaze+ \cite{li2018eye} offers approximately 10,000 samples of 106 non-scripted daily activities that occur in a kitchen and provides researchers with a first-person perspective \cite{yan2015egocentric,lu2019deep}. 

\textbf{UCF101}. The UCF101 \cite{soomro2012ucf101} dataset comprises 13,320 videos with 101 classes in the wild. Due to its limited size, video models trained on UCF101 may lean towards overfitting and makes UCF101 suitable for measuring the effectiveness of video data augmentations.

\subsection{Implementation Details}
\textbf{Baseline model}. We evaluate the effectiveness of SV-Mix on both CNN and tranformer models. Among CNN models, we choose one 2D CNN based model, i.e. TSM \cite{lin2019tsm} and one decomposed 3D CNN based model, i.e. R(2+1)D \cite{tran2018closer} because the core modules, i.e., temporal shift and 1D temporal convolstion are widely adopted as key components to construct other advanced CNN video models. We also demonstrate the merits of SV-Mix on sophisticated transformer model, e.g., ViViT \cite{arnab2021vivit}, VideoSwin \cite{liu2021video}, MViTv2 \cite{li2022mvitv2} and Uniformer \cite{li2021uniformer}. We adopt ResNet-50 and small version as the backbone for CNN and tranformer models respectively.

\begin{table*}[]
\caption{Performance comparisons with different action recognition frameworks on Sth-Sth V1$\&$V2 and Mini-Kinetics datasets.}
\centering
\small
\begin{tabular}{lcclcclcc}
\hline
\multirow{2}{*}{Method} & \multicolumn{2}{c}{Sth-Sth V1} &   & \multicolumn{2}{c}{Sth-Sth V2}  &   & \multicolumn{2}{c}{Mini-Kinetics} \\ \cline{2-3} \cline{5-6} \cline{8-9}
                        & Acc 1.(\%)  & $\Delta$Acc 1.(\%) &  & Acc 1.(\%)  & $\Delta$Acc 1.(\%) &  & Acc 1.(\%)  & $\Delta$Acc 1.(\%)  \\ \hline
TSM & 45.5  &\multirow{2}{*}{+1.7} & &  59.3  & \multirow{2}{*}{+1.0} & &  75.9  & \multirow{2}{*}{+0.7}          \\
TSM+SV-Mix & \textbf{47.2}  &  &  & \textbf{60.3}  &   &   & \textbf{76.6}    \\ \hline
R(2+1)D & 45.9  & \multirow{2}{*}{+0.8}   &  & 58.9    &  \multirow{2}{*}{+1.4}   &  & 75.5   &  \multirow{2}{*}{+0.6} \\
R(2+1)D+SV-Mix  &  \textbf{46.7}  &  &  & \textbf{60.3}  & &  &  \textbf{76.1}  &
\\ \hline
MViTv2 & 57.0  & \multirow{2}{*}{+0.9} &  & 67.4   & \multirow{2}{*}{+1.2} &  & 79.3   & \multirow{2}{*}{+0.2} \\
MViTv2+SV-Mix & \textbf{57.9} &  &  & \textbf{68.6} &  &  & \textbf{79.5} &  \\ \hline

Uniformer & 56.7  & \multirow{2}{*}{+0.7} &  &  67.7  & \multirow{2}{*}{+1.1} &  &  79.1  & \multirow{2}{*}{+0.3} \\
Uniformer+SV-Mix & \textbf{57.4} &  &  & \textbf{68.8} &  &  & \textbf{79.4} &  \\ \hline
\end{tabular}
\label{sthsth}
\vspace{-0.3cm}
\end{table*}

\begin{table*}[]
\caption{Performance comparisons on relatively small-scale datasets including UCF101, Diving48 and EGTEA GAZE+.}
\small
\centering
\label{other}
\begin{tabular}{lcclcclcc}
\hline
\multirow{2}{*}{Method}&\multicolumn{2}{c}{UCF101}&  & \multicolumn{2}{c}{Diving48}&  & \multicolumn{2}{c}{EGTEA GAZE+}                                 \\ \cline{2-3} \cline{5-6} \cline{8-9} 
             & Acc 1.(\%)   & $\Delta$ Acc 1.(\%)  &  & Acc 1.(\%)   & $\Delta$Acc 1.(\%)   &  & Acc 1.(\%)   &$\Delta$Acc 1.(\%) \\ \hline
TSM          &   85.2        &\multirow{2}{*}{+3.2}     &  &  77.6 &\multirow{2}{*}{+2.6}     &  &  63.5  & \multirow{2}{*}{+2.0}   \\
TSM+SV-Mix   &  \textbf{88.4}&                          &  &\textbf{80.2}&                      &  &\textbf{65.5}&                    \\ \hline 

ViViT       &     87.3      &\multirow{2}{*}{+1.0}&  & 70.0 &\multirow{2}{*}{+6.2}&  & 57.3  & \multirow{2}{*}{+4.8}  \\
ViViT+SV-Mix&\textbf{88.3}&                       &  &\textbf{76.2}&      &  &\textbf{62.1}&   \\ \hline

MViTv2       &90.0           &\multirow{2}{*}{+2.2}&  &  80.7&\multirow{2}{*}{+3.1}&  &  66.5 & \multirow{2}{*}{+1.3}  \\
MViTv2+SV-Mix&\textbf{92.2}&                       &  &\textbf{83.8}&      &  &\textbf{67.8}&   \\ \hline

VideoSwin       &     93.6      &\multirow{2}{*}{+3.0}&  & 78.7 &\multirow{2}{*}{+4.1}&  & 67.0  & \multirow{2}{*}{+1.8}  \\
VideoSwin+SV-Mix&\textbf{96.6}&                       &  &\textbf{82.8}&      &  &\textbf{68.8}&   \\ \hline

Uniformer       &93.2           &\multirow{2}{*}{+3.9}&  &  83.1&\multirow{2}{*}{+1.9}&  &  69.7 & \multirow{2}{*}{+2.2}  \\
Uniformer+SV-Mix&\textbf{97.1}&                       &  &\textbf{85.0}&      &  &\textbf{71.9}&   \\ \hline
\end{tabular}
\vspace{-0.3cm}
\end{table*}

\textbf{Training}. Following the common setting \cite{wang2016temporal}, we uniformly sample 8 or 16 frames from input videos for all datasets and use spatial size of $224\times 224$. Data augmentations used by baseline model such as random scaling (for CNN models) and RandAugment \cite{cubuk2020randaugment} (for transformer models) are also adopted, unless otherwise statement. 
For CNN models, we train the network via SGD optimizer and we set the learning rate (lr) as $0.01 \times \frac{batchsize}{32}$. The total training epoch is set as 50 for all datasets. At epoch 20, 40, we decay lr by multiplying 0.1. The dropout ratio is set as 0.5. 
As for transformer models, we adopt AdamW \cite{loshchilov2017fixing} instead of SGD and set lr as $2e-4 \times \frac{batchsize}{32}$. We train transformer models for 60 epochs for Something-Something datasets and 50 epochs for other datasets. Cosine learning rate schedule is adopted with 5 warmup epochs.

\textbf{Inference}. 
We sample 8 frames per video for CNN models and 16 frames for tranformer models. We utilize 224 $\times$ 224 central crop and 1 clip $\times$ 1 crop for testing CNN models except on UCF101, where we using 256 $\times$ 256 $\times$ 2 clip $\times$ 3 crop. We use 224 $\times$ 224 central crop to test transformer models, test views is set as 3 crops $\times$ 1 clip on Something-Something, 2 clip $\times$ 3 crop on UCF101 and 1 clip $\times$ 1 crop for others

\subsection{The Effectiveness of SV-Mix (RQ1)}
\textbf{Something-Something and Mini-Kinetics}. We empirically evaluate the effectiveness of adopting SV-Mix on various video models using the Something-Something dataset in Table \ref{sthsth}. 
It can be observed that our proposed SV-Mix achieves consistent performance improvement across various datasets on both CNN and transformer models.
Particularly, SV-Mix improves the performance of TSM by 1.7\% and 1.0\% on Something-Something V1 and V2, it also boosts the accuracy on Mini-Kinetics by 0.7\%. 
SV-Mix in R(2+1)D yields a 0.8\% boost on Something-Something V1, rising to 1.4\% on V2, with a 0.6\% gain in Mini-Kinetics.
When adopted to MViTv2, our SV-Mix still brings improvement by 0.9\% on Something-Something V1, 1.13\% on V2, and a 0.2\% boost in Mini-Kinetics. 
Adopting SV-Mix in Uniformer, we achieve performance gains as 0.5\% in both Something-Something V1 \& V2 and 0.3\% in Mini-Kinetics.

\textbf{Other Datasets}. We conducted further evaluations of SV-Mix on datasets with relatively small sizes, including UCF101, Diving48, and EGTEA Gaze+ in Table \ref{other}. Compared to larger datasets such as Something-Something V1\&V2, these datasets tend to suffer from more severe overfitting problems. As a result, the performance gains of SV-Mix are more significant on these datasets. Specifically, SV-Mix boosts the performance of TSM by 3.2\%, 2.6\% , 2.0\% on UCF101, Diving48 and EGTEA GAZE+, respectively. On more sophisticated transformer models, the improvements are still significant. Notably, ViVit, which only conducts temporal reasoning on the CLS token of each frame, shows relatively poorer performance, especially for Diving48 and EGTEA GAZE+, where temporal reasoning capabilities are more critical. In these cases, SV-Mix achieved the most significant performance improvements, with gains of +6.2\% and +4.8\%, respectively.

\subsection{Comparison with Other Augmentation (RQ2)}

\begin{table*}[]
\caption{Performance comparisons between SV-Mix and other data augmentation strategies on UCF101 and Sth-Sth V1 datasets.}
\small
\centering
\label{augment_compare}
\begin{tabular}{lcclcc}
\toprule[1pt]
\multirow{2}{*}{Model w/ Aug} & \multicolumn{2}{c}{UCF101}                                                   &  & \multicolumn{2}{c}{Sth-Sth V1}                                   \\ \cline{2-3} \cline{5-6} 
                              & \multicolumn{1}{c}{Acc 1.(\%)}  & $\Delta$Acc 1.(\%) &  & \multicolumn{1}{c}{Acc 1.(\%)}  & $\Delta$Acc 1.(\%) \\ \hline
TSM                                                  & 85.2         &      - &  &  45.5    &  -        \\ \hline
+Mixup \cite{zhangmixup}                             & 84.7         &   -0.5 &  &   44.6   &  -0.9          \\
+Cutmix\cite{yun2019cutmix} (VideoMix \cite{yun2020videomix})& 86.9         &   +1.7 &  &   45.7   &  +0.2          \\
+Cutmix\&Mixup                                       & 87.0         &   +1.8 &  &   45.4   &  -0.1     \\
+Cutout \cite{devries2017cutout}                     & -            &   -    &  &   44.7   &  -0.8     \\
+Augmix \cite{hendrycksaugmix}                       & -            &   -    &  &   46.2   &  +0.7     \\
+Saliency Grafting \cite{park2022saliency}           & 87.8         &  +2.6  &  &   46.4   &  +0.9     \\
+RandAug \cite{cubuk2020randaugment}                 & 87.5         &  +2.3  &  &   -      &  -        \\ \hline
+SV-Mix                                              &\textbf{88.4} &\textbf{+3.2}&  &\textbf{47.2}&\textbf{+1.7} \\
+SV-Mix+RandAug                                      &\textbf{89.6} &\textbf{+4.4}&  &     -    &   -        \\  \hline \hline 
MViTv2                                               & 90.0         &   -    &  &  57.2   &  -         \\\hline
+Cutmix\&Mixup                                       & 91.6         &  +1.6   &  &  57.0   &  -0.2      \\
+Saliency Grafting \cite{park2022saliency}           & 89.8      &  -0.2   &  &  -   &  -     \\
+TransMix \cite{chen2022transmix}                    & 91.0      &  +1.0    &  & 57.0 & -0.2        \\
+SV-Mix                                              &\textbf{92.2} &\textbf{+2.2}&  &\textbf{57.9}&\textbf{+0.7}   \\ \hline \hline
\end{tabular}
\vspace{-0.3cm}
\end{table*}

We compare our proposed SV-Mix with advanced data augmentation methods, as these methods are design for image augmentation, we simply expand them into video version by conducting the same augmentation in all frames, as most works do \cite{fan2020pyslowfast,2020mmaction2,li2022mvitv2,liu2021video}. We conduct comparison using TSM and MViTv2 on UCF101 and Something-Something V1. As shown in Table \ref{augment_compare}, simply deleting (i.e. Cutout \cite{devries2017cutout}) or exchanging (i.e. Cutmix/VideoMix \cite{yun2019cutmix,yun2020videomix} \footnote{It is worth notice that VideoMix \cite{yun2020videomix} explore the adaption of Cutmix \cite{yun2019cutmix} in video recognition and propose to cut the same regions for all frames in the video. Because this strategy simply inflates Cutmix along the temporal dimension, we refer to VideoMix as Cutmix in the following part of this paper as popular projects do \cite{fan2020pyslowfast,2020mmaction2}.}) a random spatial region bring little improvement or even decrease the recognition accuracy, we infer it may due to core motion area missing cause by these two methods. Mixup \cite{zhangmixup} which blurs the whole frame also brings no consistent enhancement on recognition accuracy. Augmix \cite{hendrycksaugmix} which mixes several augmented views of one video sample and RandAugment \cite{cubuk2020randaugment} which randomly selects augmentations from a pre-defined augmentation set significantly boost the performance as they enrich the training diversity and maintain the core motion pattern. 
Advanced Saliency Grafting \cite{park2022saliency} and TransMix \cite{chen2022transmix} demonstrate significance because of they respectively reconstruct mixed samples and mixed labels.
Compared with methods with sample mixing \cite{yun2019cutmix,zhangmixup}, Cutout \cite{devries2017cutout} and advanced mixing \cite{park2022saliency, chen2022transmix}, our SV-Mix perform much better consistently because of the motion pattern selection capability (spatial/temporal selective module) and better sample diversity (random volume partition).

\subsection{Ablation Study (RQ3)}

\begin{table}[]

\caption{Ablation study of spatial selective module and temporal selective module in SV-Mix on Sth-Sth V1 dataset.}
\small
\centering
\label{ablation}
\begin{tabular}{lcc}
\hline
\multirow{2}{*}{Model w/ Mixups} & \multicolumn{2}{c}{Sth-Sth V1}           \\ \cline{2-3}
                                 & Acc 1.(\%)  & $\Delta$Acc 1.(\%) \\ \hline
TSM                              & 45.5&  -     \\ \hline\hline

+Cutmix                          & 45.7&  +0.2     \\
+Spa. Select                     &\textbf{47.0}&\textbf{+1.5}    \\ \hline
+Mixup                           & 44.6&  -0.9     \\
+Temp. Select                    &\textbf{46.6}&\textbf{+1.1}     \\ \hline
+Mixup\&Cutmix                   & 45.4&  -0.1     \\
+Temp.\&Spa. Select              &\textbf{47.2}&\textbf{+1.7}     \\ \hline
\end{tabular}
\end{table}



\begin{table}[]
\caption{Performance comparison between disentangled training and entangled training on Sth-Sth V1 dataset.}
\small
\centering
\label{tab:en_vs_disen}
\begin{tabular}{l|cc|c}

\hline
Mix Module                          & En       & DisEn    & Top1(\%) \\ \hline
TSM                                 & ---      & ---      & 45.5     \\ \hline \hline
\multirow{2}{*}{+Spa. Select}       &\ding{52} &          &45.6     \\
                                    &          &\ding{52} & \textbf{47.0}     \\ \hline
\multirow{2}{*}{+Temp. Select}      &\ding{52} &          & 45.2     \\
                                    &          &\ding{52} & \textbf{46.6}     \\ \hline
\multirow{2}{*}{+Temp\&Spa. Select} &\ding{52} &          & 45.8     \\
                                    &          &\ding{52} & \textbf{47.2}     \\ \hline
\end{tabular}
\vspace{-0.3cm}
\end{table}

\begin{figure}[t]
  \centering
  \includegraphics[width=0.9\linewidth]{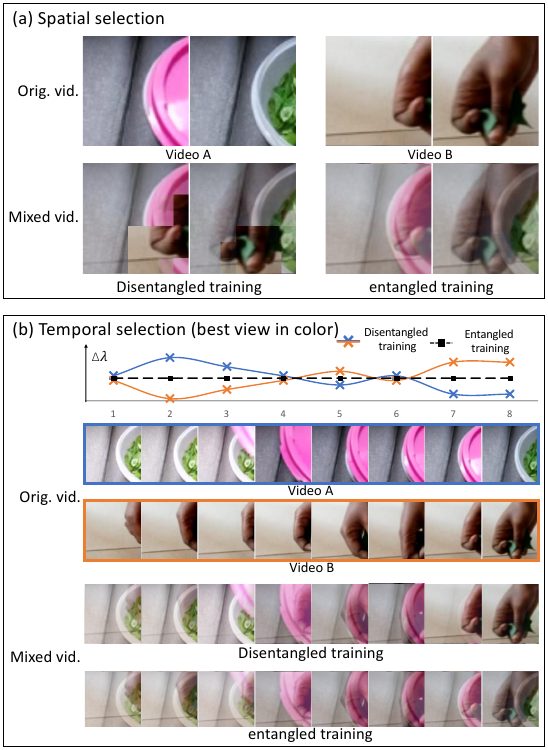}
    \caption{Instance visualization of mixing two videos labeled as ``\emph{Pretending to close sth}'' (Video A) and  ``\emph{Tearing sth into two pieces}'' (Video B). We compare mix videos generated by spatial selective module and temporal selective module under disentangled training and entangled training to verify the importance of training disentanglement. Both spatial and temporal selective modules fail to capture the informative spatial/temporal volumes and mix video samples in a uniform manner.}
  \label{fig:disen}
\end{figure}

\begin{table}[]
\vspace{-0.2cm}
\caption{Effectiveness of $\lambda$ embedding and $\mathcal{L}_\mathcal{M}$.}
\small
\centering
\label{tab:lamda}
\begin{tabular}{l|cc|c}

Mix Module                          &$\lambda$ Em.&$\mathcal{L}_\mathcal{M}$& Top1(\%) \\ \midrule[1pt] 
TSM                                 & ---         & ---                     & 45.5     \\ \hline \hline
\multirow{4}{*}{+SV-Mix}            & \ding{56}   & \ding{56}               & 45.1     \\
                                    & \ding{52}   &  \ding{56}               & 46.9     \\ 
                                    & \ding{56}   & \ding{52}                & 45.9     \\
                                    & \ding{52}   &\ding{52}                 & \textbf{47.2}     \\  \hline

\end{tabular}
\vspace{-0.3cm}
\end{table}

\begin{table}[]
\caption{Performance comparison of different ensemble strategies and parameter sharing settings of spatial/temporal selective modules.}
\small
\centering
\label{tab:ensemble}
\begin{tabular}{l|cc|c}

Mix Module                          &Prob. En.     &Para. Share              & Top1(\%) \\ \midrule[1pt] 
TSM                                 & ---          & ---                     & 45.5     \\ \hline \hline
\multirow{4}{*}{+SV-Mix}            & \ding{56}    & \ding{56}               & 45.9 \\
                                    & \ding{52}    & \ding{56}               & 46.2     \\ 
                                    & \ding{56}    & \ding{52}               & 47.0 \\
                                    & \ding{52}    &\ding{52}                & \textbf{47.2}     \\  \hline

\end{tabular}
\vspace{-0.5cm}
\end{table}

\begin{figure}[t]
  \centering
  \includegraphics[width=0.85\linewidth]{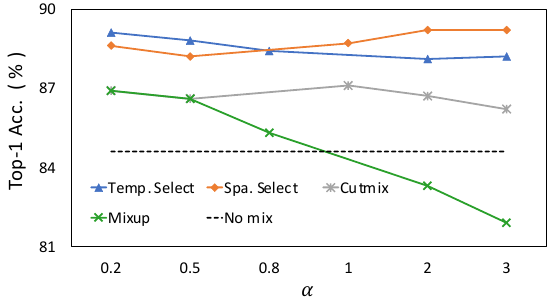}
    \vspace{-0.3cm}
    \caption{The accuracy curve of SV-Mix with different $\alpha$ for the beta distribution $\lambda\sim Beta(\alpha,\alpha)$ on UCF101 dataset.}
  \label{fig:beta}
  \vspace{-0.5cm}
\end{figure}



In this subsection, we demonstrate ablation studies of SV-Mix to verify the merits of our design choose, including the effectiveness of spatial/temporal selective module, disentangled training and the robustness under hyperparameter changing.

\textbf{Spatial/temporal selective module.} For better understanding about how the components of SV-Mix influence the performance of action recognition, we ablate spatial selective module and temporal selective module in Table \ref{ablation}, as well as compare them with vanilla TSM model and their counterparts, i.e. CutMix \cite{yun2019cutmix} and Mixup \cite{zhangmixup}. As shown in Table \ref{ablation}, individual spatial selective module or temporal selective module already enhances the performance of TSM model (+1.52\% and +1.10\% respectively) which significantly outperform their non-parameter counterparts. Further ensembling spatial selective module and temporal selective module by random switching boosts the performance improvement to a higher level (+1.75\%).

\textbf{Disentangled training.} We compare the proposed disentangled training pipeline with the intuitive entangled training pipeline using something-something V1 dataset. 
As shown in Table \ref{tab:en_vs_disen}, disentangled training consistently outperforms intuitive entangled training under different mix module settings by substantial margins (+1.4\%). To illustrate the contrast between the two training strategies more vividly, we provide comparison between mixed video samples generated by spatial/temporal selective module trained under these two training strategies. As shown in Figure \ref{fig:disen} (a), trained by disentangled pipeline, spatial selective module keeps the fingertip actions areas that are highly correlated with the label \emph{``Tearing sth into two pieces''} in video B for mixed sample generation. As a contrast spatial selective module fall into sub-optimal where videos are uniformly mixed across spatial regions. In temporal volume selection, similar phenomenon is observed. Specifically, as illustrated in the first row of Figure \ref{fig:disen} (b), when trained by disentangled pipeline, temporal selective module is able to identify the first few frames in video A and the last few frames in video B as the informative frames which should be assigned with higher weights in the mixed video. When optimized by entangled training strategy, temporal selective module assigns similar weights for all frames and then fails to emphasize the informative frames in the mixed sample.

\textbf{$\lambda$ embedding $\&$ $\mathcal{L}_\mathcal{M}$}. By embedding $\lambda$ into the volume features through concatenation, SV-Mix is capable to control the mixing proportion of video samples. Additionally, $\mathcal{L}_\mathcal{M}$ provides explicitly guidance for SV-Mix to build correlation between $\lambda$ and mixed training samples. The effectiveness of $\lambda$ embedding and $\mathcal{L}_\mathcal{M}$ is illustrated in Table \ref{tab:lamda}. In particular, in the absence of $\lambda$ embedding and $\mathcal{L}_\mathcal{M}$, SV-Mix fails to improve the recognition performance (45.5$\%$ $\rightarrow$ 45.1$\%$). 
With $\lambda$ embedded in the input, SV-Mix boosts the TSM to reach 46.9$\%$ and the introduction $\mathcal{L}_\mathcal{M}$ further enhances the model to 47.2$\%$.

\textbf{Ensemble strategies and parameter sharing}. We conduct performance comparison of different ensemble strategies and parameter sharing of spatial and temporal selective modules. As illustrated in Table \ref{tab:ensemble}, using shared parameters of spatial and temporal selective modules consistently outperforms unshared parameters settings. In addition, probabilistic ensemble (Eq \ref{eq:pe}) of spatial and temporal selective modules provide slight improvements over average ensemble (Eq \ref{eq:ae}).

\textbf{Distribution of $\lambda$.} We further explore the influence of different distribution of $\lambda$ by varying the $\alpha$ value which determines the distribution of $\lambda$ as $\lambda \sim Beta(\alpha,\alpha)$. The results on UCF101 dataset are shown in Figure \ref{fig:beta}. 
We evaluate the performance of spatial selective module and temporal selective module using TSM as backbone on $\alpha=\{0.2,0.5,1,2,3\}$ and $\alpha=\{0.2,0.5,0.8,2,3\}$ respectively. 
As shown in Figure \ref{fig:beta}, temporal selective module and spatial selective module demonstrate significant and robust performance improvement under different $\lambda$ distribution, especially compared with mixup \cite{zhangmixup}. 
The performance of spatial selective module decreases when $\alpha$ goes lower than 1, which may be due to the fact that when $\alpha$ is lower than 1, the spatial selective module selects smaller spatial regions that may not contain complete motion regions, leading to semantic ambiguity in the training data-label pairs. In contrast, the performance of temporal selective module improves with smaller $\alpha$. There may be two reason for this phenomenon, 1) temporal selective module select the whole frame instead of a small spatial region; 2) actions in UCF101 can be recognized using only a few frames. Notably, in other experiments, we fixed $\alpha$ as 0.8 and 1 for temporal and spatial selective module respectively, although adjusting $\alpha$ may lead to a slight improvement.



\subsection{Analysis and Visualization}

\begin{table}[]
\caption{Comparison across various model depth.}
\centering
\begin{tabular}{lccc}
\toprule
Backbone  & TSM  &                       &  TSM+SV-Mix  \\ \hline  
ResNet18  &37.5 &$\xrightarrow{+1.9 \%}$& 39.4 \\
ResNet50  & 45.5 &$\xrightarrow{+1.7 \%}$& 47.2  \\
ResNet101  & 47.6 &$\xrightarrow{+1.1 \%}$& 48.7 \\
ResNet152  & 48.8 &$\xrightarrow{+1.0 \%}$& 49.8 \\
\hline
\end{tabular}
\label{tab:depth}
\end{table}

\begin{figure}[t]
  \centering
  \includegraphics[width=\linewidth]{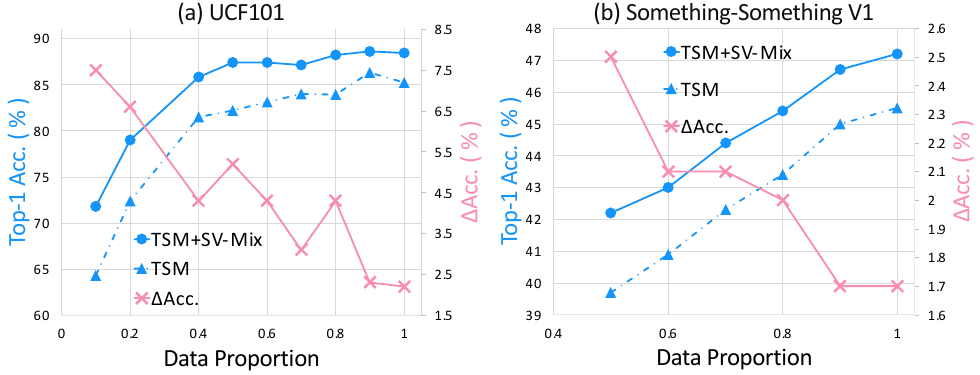}
  \vspace{-0.5cm}
  \caption{The overfitting intensified by the decrease of amount of training data, making the benefit of SV-Mix more significant.}
  \label{overfitting}
  \vspace{-0.5cm}
\end{figure}


\begin{figure*}[t]
  \centering
  \includegraphics[width=\linewidth]{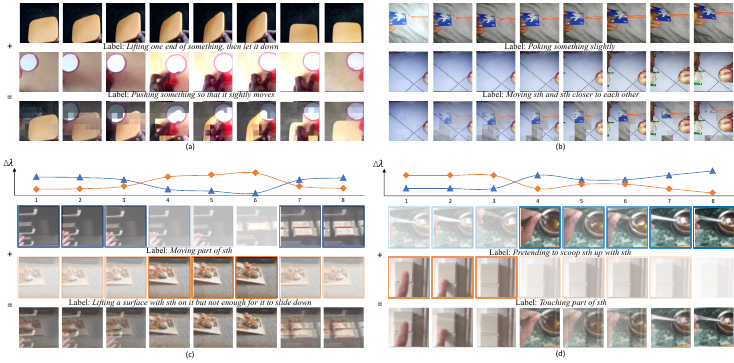}
      \vspace{-0.5cm}
  \caption{Examples of augmented video by SV-Mix. (a) and (b) are generated by spatial selection, (c) and (d) are generated by temporal selection.}
  \label{spa_tem_vis}
  \vspace{-0.2cm}
\end{figure*}


\textbf{Model depth.} 
We compared the performance of SV-Mix on models of varying depths, using accuracy on the Something-Something V1 as the metric. As shown in Table \ref{tab:depth}, as the backbone model deepens from ResNet18 to ResNet152, the baseline performance shows an upward trend. At the same time, SV-Mix brings significant performance improvements. However, the performance improvement by SV-Mix decreases as the baseline performance improves, likely due to the increased difficulty of further enhancing performance at higher baseline levels.

\textbf{Impact of dataset scale.} 
To demonstrate the role of SV-Mix in preventing overfitting, we artificially reduced the amount of training data to intensify overfitting and observed the changes of the effects of SV-Mix.
Specifically, we kept the test set of the dataset unchanged and increased overfitting by proportionally reducing the size of the training set. As shown in Figure \ref{overfitting}, as the size of the training set decreases, the model’s performance declines, while SV-Mix achieves greater performance improvements. This demonstrates the significance of SV-Mix in preventing overfitting. 

\textbf{Generated samples.} To demonstrate how SV-Mix works, we provide examples generated by the spatial selective module and temporal selective module. Figure \ref{spa_tem_vis} (a) shows the mixing process of two videos with label \emph{``Lifting one end of sth, then let it down''} (upper row) and \emph{``Pushing sth so that it sightly moves''} (middle row) respectively. Spatial selective module successfully selects patches that contain the interaction of \emph{``hand''} and \emph{``mirror''} in 4th $\sim$ 7th frames.
Figure \ref{spa_tem_vis} (b) demonstrates mixing a video with label \emph{``Moving something and something closer to each other"} (middle row) into a video labeled \emph{``Poking something slightly''} (upper row), spatial selective module captures the region importance of both videos and maintains salient patches in the mixed video.


Unlike spatial selective module, our temporal selective module arranges whole frames and maintains the spatial pattern unchanged. We visualize the dynamic weights it assigns for different frames as well as the mixed video samples. Figure \ref{spa_tem_vis} (c) demonstrates the mixing process of a video labeled \emph{``Moving part of sth''} (upper row) and a video labeled \emph{``Lifting a surface with sth on it but not enough for it to slide down''} (middle row). Our temporal selective module selects the 1st $\sim$ 3rd frames and 7th $\sim$ 8th frames from the \emph{``Moving part of something''} video because these frames show the start of the action (hand on the handle, pretend to push) and the end of the action (opened drawer), respectively. For the \emph{``Lifting a surface with sth on it but not enough for it to slide down''} video, our temporal selective module correctly selects the frames that capture the process of the surface being lifted.
While in Figure \ref{spa_tem_vis} (d), a video labeled \emph{``Touching part of sth''} (middle row) is mixed into a video with label \emph{``Pretending to scoop sth up with sth''} (upper row), our temporal selective module assigns high weights for the first 3 frames of video in middle row and low weights for the rest of this video because the motion information is concentrated in the first 3 frames.



\begin{figure}[t]
  \centering
  \includegraphics[width=0.9\linewidth]{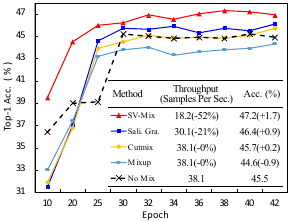}
  \vspace{-0.2cm}
  \caption{
  Throughput and training state comparison between different data augmentations on Sth-Sth V1 dataset. SV-Mix achieved the best performance improvement at the cost of training efficiency. Meanwhile SV-Mix provides better convergence speed at early epochs.
  }
  \label{state}
  \vspace{-0.5cm}
\end{figure}

\textbf{Training cost and training state.} We measure the training cost of our SV-Mix by the training throughput using a single 3090 GPU in Figure \ref{state}. In which, we also explore the model's training state under different data augmentation methods as an Accuracy-Epoch curve.
SV-Mix results in a significant decrease in efficiency, with throughput dropping by more than 50\% and achieves the best performance enhancement. 
The Accuracy-Epoch curve indicates that simple Cutmix and Mixup methods slow down the model convergence (i.e., 10$\sim$20 epochs) and stabilize the training process (i.e., 20$\sim$25 epochs), but they don't bring significant performance improvements. In contrast, our SV-Mix exhibit no convergence issues and even accelerate the model training at an early epochs. This suggests a potential efficient training protocol for enhancing the effectiveness of SV-Mix, which is considered in our future works.


\section{Conclusion and limitation}
We have presented SV-Mix augmentation, which provides a learnable data augmentation strategy for video action recognition. Particularly, we formulate the learnable video mixing process as the attention mechanism across volumes from two videos. 
The volumes with the most distinctive content compared with another video are treated as informative volumes, which should be maintained in the mixed video. To materialize our idea, we devise spatial selective module and temporal selective module to seek the valuable volumes on patch level and frame level, respectively. By randomly choosing one of the two modules, SV-Mix can produce both spatially mixed video and temporally mixed video. The modules in SV-Mix are jointly optimized with the subsequent action recognition framework in a novelly designed disentangled manner. The results of SV-Mix on five action recognition datasets demonstrate a consistent improvements across different benchmarks. Furthermore, as shown in the experiments with different recognition frameworks, SV-Mix demonstrates good potential to benefit a large range of neural networks from 2D CNNs, 3D CNNs to video transformers.

This study investigates the effectiveness of volume selection in action recognition. However, there remains an outstanding issue regarding the efficiency of SV-Mix. The disentangled training pipeline of SV-Mix necessitates multiple forward propagations of the backbone model, which results in longer training time. Efficient volume selection based video data augmentation is left for our future research and improvement.


\bibliographystyle{IEEEtran}
\bibliography{refer}

\vfill

\end{document}